%% file: main.tex
\documentclass[final]{cvpr}

\usepackage{times}
\usepackage{epsfig}
\usepackage{graphicx}
\usepackage{amsmath}
\usepackage{amssymb}
\usepackage{multirow}
\usepackage{subfigure}
\usepackage{nidanfloat}
\usepackage{diagbox}

\usepackage[dvipsnames]{xcolor}

\makeatletter\renewcommand\paragraph{\@startsection{paragraph}{4}{\z@}
  {.5em \@plus1ex \@minus.2ex}{-.5em}{\normalfont\normalsize\bfseries}}\makeatother

\usepackage[pagebackref=true,breaklinks=true,colorlinks,bookmarks=false]{hyperref}

\begin{document}

\title{Simple Copy-Paste is a Strong Data Augmentation Method\\ for Instance Segmentation}

\author{
Golnaz Ghiasi\thanks{Equal contribution. Correspondence to: \href{mailto:golnazg@google.com}{golnazg@google.com}.}~~$^{1}$ \quad Yin Cui\footnotemark[1]~~$^{1}$ \quad Aravind Srinivas\footnotemark[1]~~\thanks{Work done during an internship at Google Research.}~~$^{1,2}$ \\
Rui Qian\footnotemark[2]~~$^{1,3}$ \quad Tsung-Yi Lin$^{1}$ \quad  Ekin D. Cubuk$^{1}$ 
\quad Quoc V. Le$^{1}$ \quad Barret Zoph$^{1}$\\
\\
$^1$Google Research, Brain Team \quad $^2$ UC Berkeley \quad $^3$ Cornell University\\
}

\maketitle

\input{abstract}
\input{introduction}

\input{related}

\begin{figure*}[t]
    \centering
    \subfigure[\textbf{Standard Scale Jittering (SSJ)}]{\includegraphics[width=0.37\textwidth ]{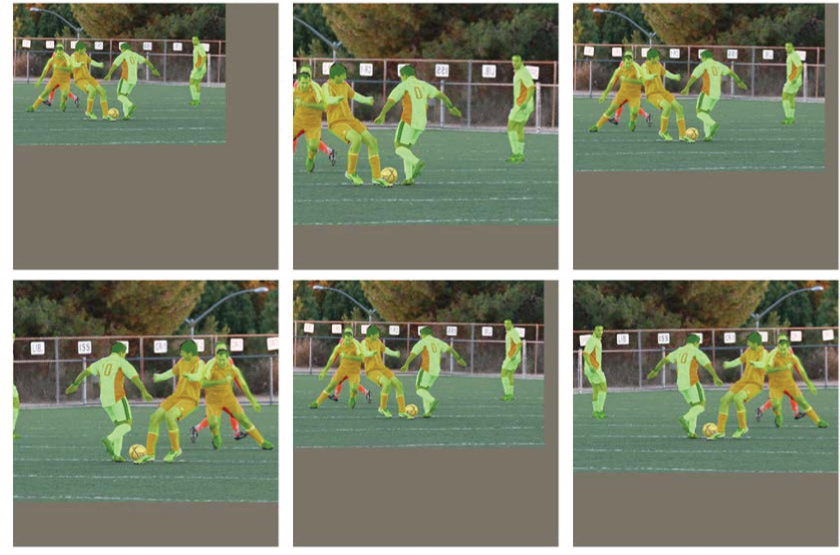}} \hspace{16mm}
    \subfigure[\textbf{Large Scale Jittering (LSJ)}]{\includegraphics[width=0.37\textwidth]{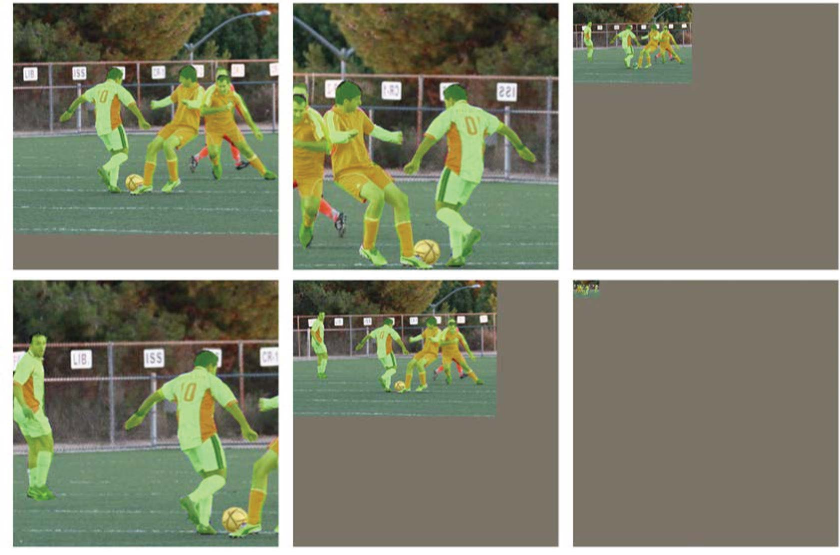}}
    \caption{Notation and visualization of the two scale jittering augmentation methods used throughout the paper. Standard Scale Jittering (SSJ) resizes and crops an image with a resize range of 0.8 to 1.25 of the original image size. The resize range in Large Scale Jittering (LSJ) is from 0.1 to 2.0 of the original image size. If images are made smaller than their original size, then the images are padded with gray pixel values. Both scale jittering methods also use horizontal flips.}
    \label{fig:scale_jittering}
\vspace{-2mm}
\end{figure*}

\section{Method}
\label{sec:method}

Our approach for generating new data using Copy-Paste is very simple. We randomly select two images and apply random scale jittering and random horizontal flipping on each of them. Then we select a random subset of objects from one of the images and paste them onto the other image. Lastly, we adjust the ground-truth annotations accordingly: we remove fully occluded objects and update the masks and bounding boxes of partially occluded objects.

Unlike~\cite{fang2019instaboost, dvornik2018modeling}, we do not model the surrounding context and, as a result, generated images can look very different from real images in terms of co-occurrences of objects or related scales of objects. For example, giraffes and soccer players with very different scales can appear next to each other (see Figure~\ref{fig:method}).

\paragraph{Blending Pasted Objects.}
For composing new objects into an image, we compute the binary mask ($\alpha$) of pasted objects using ground-truth annotations and compute the new image as  $ I_1 \times \alpha + I_2 \times (1-\alpha) $ where $I_1$ is the pasted image and $I_2$ is the main image. To smooth out the edges of the pasted objects we apply a Gaussian filter to $\alpha$ similar to ``blending'' in~\cite{dwibedi2017cut}. But unlike~\cite{dwibedi2017cut}, we also found that simply composing without any blending has similar performance.

\paragraph{Large Scale Jittering.}
We use two different types of augmentation methods in conjunction with Copy-Paste throughout the text: standard scale jittering (SSJ) and large scale jittering (LSJ). These methods randomly resize and crop images. See Figure~\ref{fig:scale_jittering} for a graphical illustration of the two methods. In our experiments we observe that the large scale jittering yields significant performance improvements over the standard scale jittering used in most prior works.

\paragraph{Self-training Copy-Paste.}
In addition to studying Copy-Paste on supervised data, we also experiment with it as a way of incorporating additional unlabeled images. Our self-training Copy-Paste procedure is as follows: (1) train a supervised model with Copy-Paste augmentation on labeled data, (2) generate pseudo labels on unlabeled data, (3) paste ground-truth instances into pseudo labeled and supervised labeled images and train a model on this new data.

\section{Experiments}

\subsection{Experimental Settings}

\paragraph{Architecture.} We use Mask R-CNN~\cite{he2017mask} with EfficientNet~\cite{tan2019efficientnet} or ResNet~\cite{he2016deep} as the backbone architecture. We also employ feature pyramid networks~\cite{lin2017feature} for multi-scale feature fusion. We use pyramid levels from $P_2$ to $P_6$, with an anchor size of $8\times2^l$ and 3 anchors per pixel.
Our strongest model uses Cascade R-CNN~\cite{cai2018cascade}, EfficientNet-B7 as the backbone and NAS-FPN~\cite{ghiasi2019fpn} as the feature pyramid with levels from $P_3$ to $P_7$. The anchor size is $4\times2^l$ and we have 9 anchors per pixel. Our NAS-FPN  model uses 5 repeats and we replace convolution layers with ResNet bottleneck blocks~\cite{he2016deep}.

\paragraph{Training Parameters.}
All models are trained using synchronous batch normalization~\cite{ioffe2015batch, girshick2018detectron} using a batch size of 256 and weight decay of 4e-5. We use a learning rate of 0.32 and a step learning rate decay~\cite{he2019rethinking}. At the beginning of training the learning rate is linearly increased over the first 1000 steps from 0.0032 to 0.32. We decay the learning rate at 0.9, 0.95 and 0.975 fractions of the total number of training steps. We initialize the backbone of our largest model from an ImageNet checkpoint pre-trained with self-training~\cite{xie2020self} to speed up the training. All other results are from models with random initialization unless otherwise stated. Also, we use large scale jittering augmentation for training the models unless otherwise stated. 
For all different augmentations and dataset sizes in our experiments we allow each model to train until it converges (\ie, the validation set performance no longer improves). For example, training a model from scratch with large scale jittering and Copy-Paste augmentation requires 576 epochs while training with only standard scale jittering takes 96 epochs. For the self-training experiments we double the batch size to 512 while we keep all the other hyper-parameters the same with the exception of our largest model where we retain the batch size of 256 due to memory constraints.

\paragraph{Dataset.}
We use the COCO dataset~\cite{lin2014microsoft} which has 118k training images. For self-training experiments, we use the unlabeled COCO dataset (120k images) and the Objects365 dataset~\cite{shao2019objects365} (610k images) as unlabeled images.
For transfer learning experiments, we pre-train our models on the COCO dataset and then fine-tune on the Pascal VOC dataset~\cite{pascal}. For semantic segmentation, we train our models on the \texttt{train} set (1.5k images) of the PASCAL VOC 2012 segmentation dataset. For detection, we train on the \texttt{trainval} set of PASCAL VOC 2007 and PASCAL VOC 2012.
We also benchmark Copy-Paste on LVIS v1.0 (100k training images) and report results on LVIS v1.0 \texttt{val} (20k images). LVIS has 1203 classes to simulate the long-tail distribution of classes in natural images.

\begin{figure*}[t]
    \centering
    \subfigure{\includegraphics[width=0.37\textwidth]{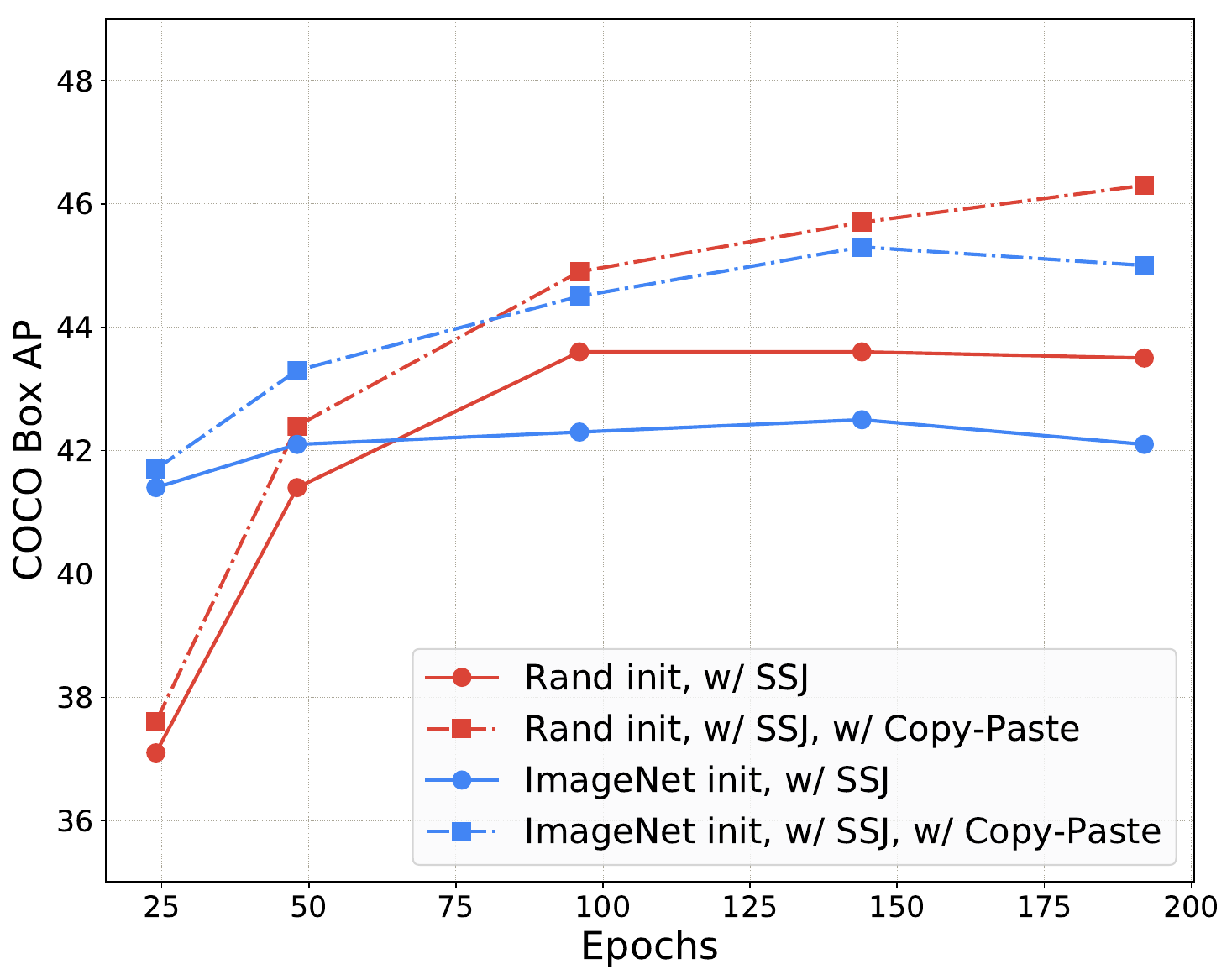}} \hspace{16mm}
    \subfigure{\includegraphics[width=0.37\textwidth]{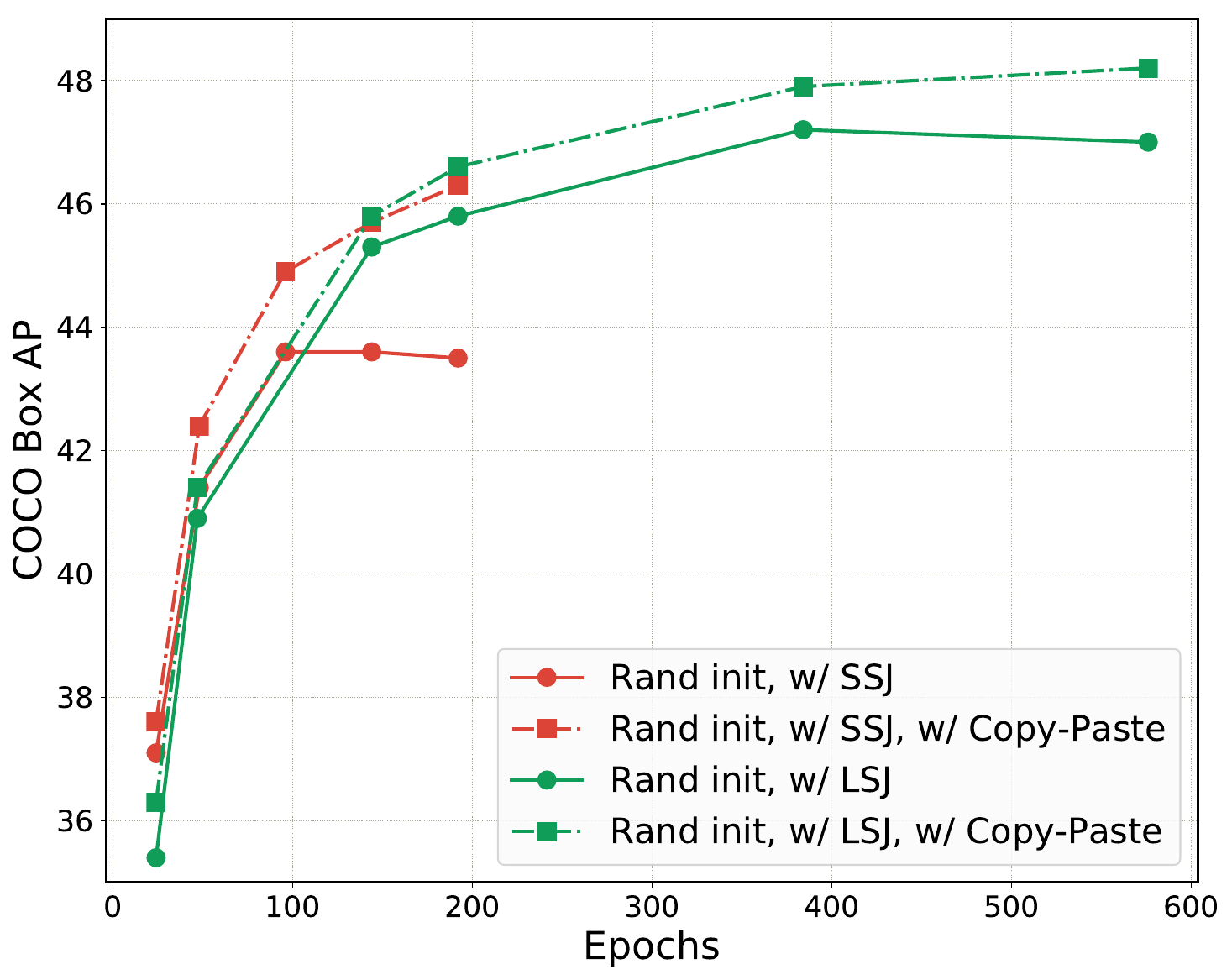}}
    \caption{Copy-Paste provides gains that are robust to training configurations. We train Mask R-CNN (ResNet-50 FPN) on 1024$\times$1024 image size for varying numbers of epochs. \textbf{Left Figure:} Copy-Paste with and without initializing the backbone by ImageNet pre-training. \textbf{Right Figure:} Copy-Paste with standard and large scale jittering. Across all of the configurations training with Copy-Paste is helpful.}
    \label{fig:vary_epochs}
\vspace{-2mm}
\end{figure*}

\subsection{Copy-Paste is robust to training configurations}
In this section we show that Copy-Paste is a strong data augmentation method that is robust across a variety of training iterations, models and training hyperparameters.

\paragraph{Robustness to backbone initialization.}
Common practice for training Mask R-CNN is to initialize the backbone with an ImageNet pre-trained checkpoint. However He~\etal~\cite{he2019rethinking} and Zoph~\etal~\cite{zoph2020rethink} show that a model trained from random initialization has similar or better performance with longer training. Training models from ImageNet pre-training with strong data-augmentation (\ie RandAugment~\cite{cubuk2020randaugment}) was shown to hurt the performance by up to 1 AP on COCO. Figure~\ref{fig:vary_epochs} (left) demonstrates that Copy-Paste is additive in both setups and we get the best result using Copy-Paste augmentation and random initialization.

\paragraph{Robustness to training schedules.}
A typical training schedule for Mask R-CNN in the literature is only 24 (2$\times$) or 36 epochs (3$\times$)~\cite{he2019rethinking, he2017mask, fang2019instaboost}. However, recent work with state-of-the-art results show that longer training is helpful in training object detection models on COCO~\cite{zoph2020rethink, tan2019efficientdet, du2019spinenet}. Figure~\ref{fig:vary_epochs} shows that we get gains from Copy-Paste for the typical training schedule of 2$\times$ or 3$\times$ and as we increase training epochs the gain increases. This shows that Copy-Paste is a very practical data augmentation since we do not need a longer training schedule to see the benefit.

\begin{figure*}[ht!]
    \centering
    \subfigure{\includegraphics[width=0.37\textwidth]{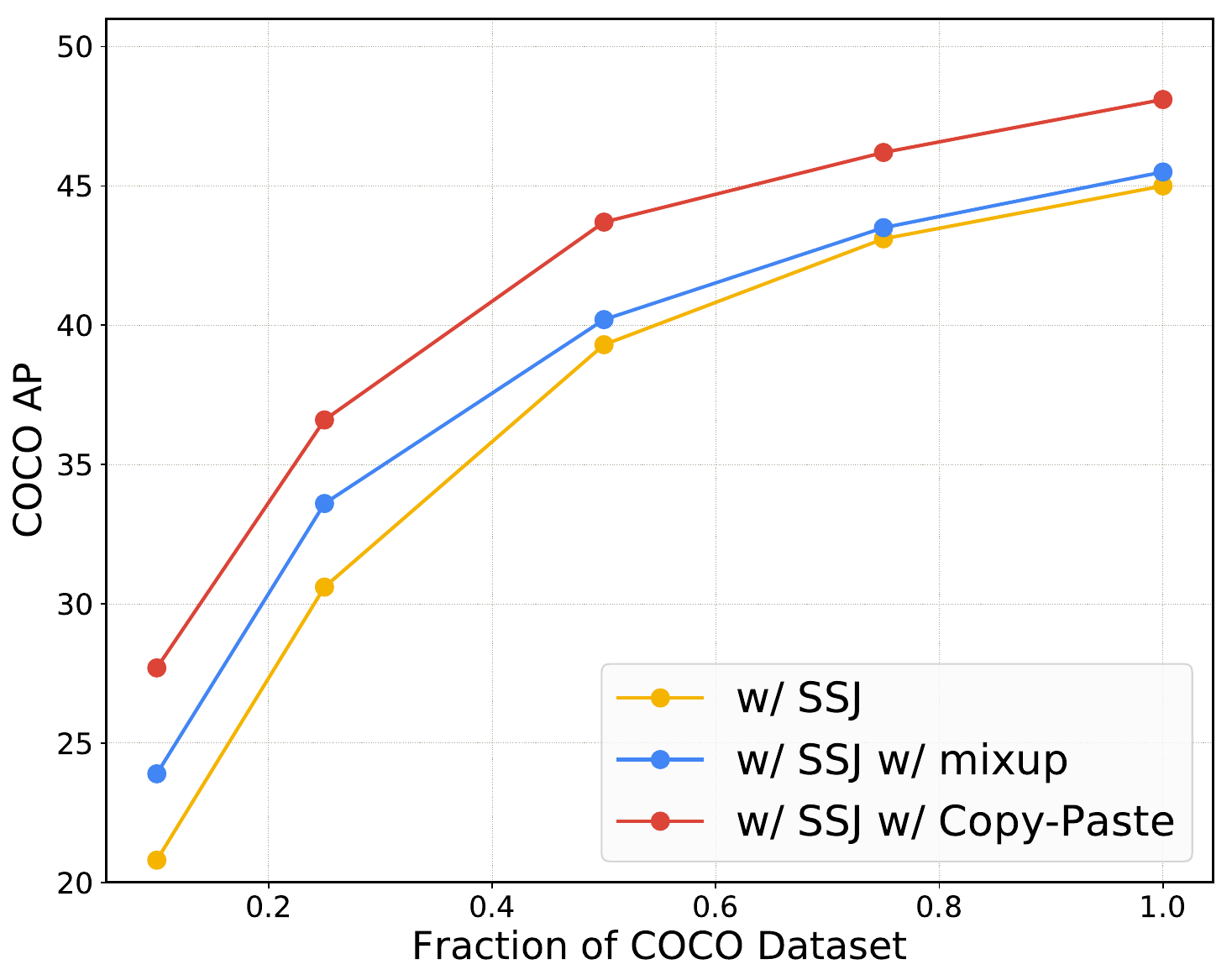}} \hspace{16mm}
    \subfigure{\includegraphics[width=0.37\textwidth]{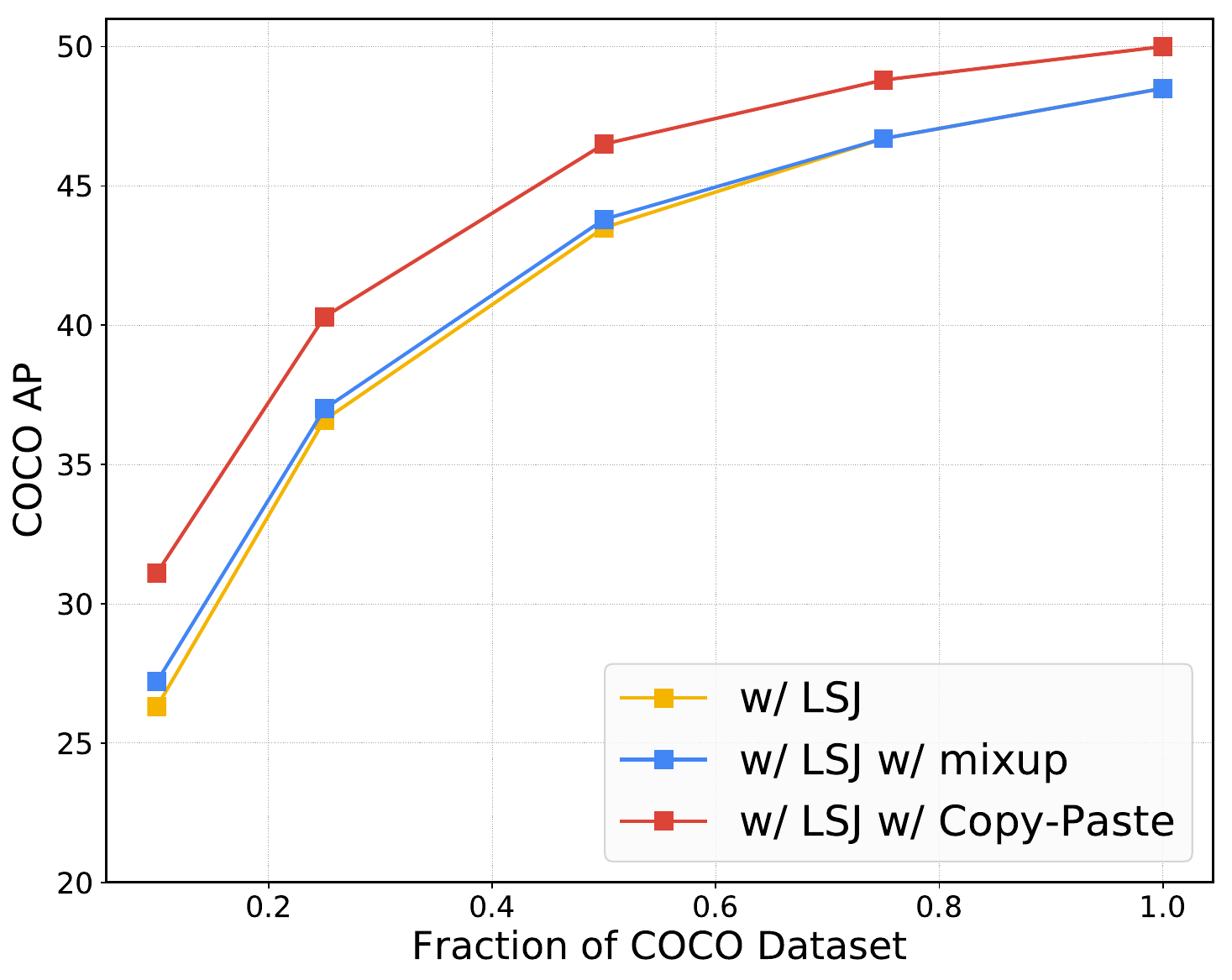}}
    \caption{Copy-Paste is additive to large scale jittering augmentation. Improvement from mixup and Copy-Paste data augmentation on top of standard scale jittering  \textbf{(Left Figure)} and large scale jittering \textbf{(Right Figure)}. All results are from training Mask R-CNN EfficientNetB7-FPN on the image size of 640$\times$640.}
    \label{fig:coco_fraction}
\vspace{-2mm}
\end{figure*}

\paragraph{Copy-Paste is additive to large scale jittering augmentation.}
Random scale jittering is a powerful data augmentation that has been used widely in training computer vision models. The standard range of scale jittering in the literature is 0.8 to 1.25~\cite{lin2017focal, he2019rethinking, cubuk2018autoaugment, fang2019instaboost}. However, augmenting data with larger scale jittering with a range of 0.1 to 2.0~\cite{tan2019efficientdet,du2019spinenet} and longer training significantly improves performance (see Figure~\ref{fig:vary_epochs}, right plot). Figure~\ref{fig:coco_fraction} demonstrates that Copy-Paste is additive to both standard and large scale jittering augmentation and we get a higher boost on top of standard scale jittering. On the other hand, as it is shown in Figure~\ref{fig:coco_fraction}, mixup~\cite{zhang2017mixup, zhang2019bag} data augmentation does not help when it is used with large scale jittering.

\paragraph{Copy-Paste works across backbone architectures and image sizes.}
Finally, we demonstrate Copy-Paste helps models with standard backbone architecture of ResNet~\cite{he2016deep} as well the more recent architecture of EfficientNet~\cite{tan2019efficientnet}. We train models with these backbones on the image size of 640$\times$640, 1024$\times$1024 or 1280$\times$1280. Table~\ref{tab:model_sizes} shows that we get significant improvements over the strong baselines trained with large scale jittering for all the models. Across 7 models with different backbones and images sizes Copy-Paste gives on average a 1.3 box AP and 0.8 mask AP improvement on top of large scale jittering. 

\begin{table}[t]
\footnotesize
\centering
\begin{tabular}{lrrr}
  \hline
  Model  & FLOPs  &  Box AP & Mask AP \\
  \hline
  Res-50 FPN (1024) & 431 B & 47.2 & 41.8 \\
  w/ Copy-Paste  & 431 B  & \textcolor{blue}{(+1.0)} 48.2 & \textcolor{blue}{(+0.6)} 42.4 \\
  \hline
  Res-101 FPN (1024) & 509 B  & 48.4 & 42.8 \\
  w/ Copy-Paste & 509 B  & \textcolor{blue}{(+1.4)} 49.8 & \textcolor{blue}{(+0.8)} 43.6 \\
  \hline
  Res-101 FPN (1280) & 693 B  & 49.1 & 43.1 \\
  w/ Copy-Paste & 693 B  & \textcolor{blue}{(+1.2)} 50.3 & \textcolor{blue}{(+1.1)} 44.2 \\
  \hline
  Eff-B7 FPN (640)  & 286 B  & 48.5 & 42.7\\
  w/ Copy-Paste & 286 B  & \textcolor{blue}{(+1.5)} 50.0 & \textcolor{blue}{(+1.0)} 43.7\\
  \hline
  Eff-B7 FPN  (1024) & 447 B  & 50.8 & 44.7\\
  w/ Copy-Paste & 447 B & \textcolor{blue}{(+1.1)} 51.9 & \textcolor{blue}{(+0.5)} 45.2\\
  \hline
  Eff-B7 FPN  (1280) & 595 B  & 51.1 & 44.8 \\
  w/ Copy-Paste & 595 B  & \textcolor{blue}{(+1.5)} 52.6 & \textcolor{blue}{(+1.1)} 45.9\\
  \hline
  Cascade Eff-B7 FPN(1280)  & 854 B & 52.9 & 45.6 \\
  w/ Copy-Paste & 854 B & \textcolor{blue}{(+1.1)} 54.0 & \textcolor{blue}{(+0.7)} 46.3\\
  \hline
\end{tabular}
\vspace{1mm}
\caption{Copy-paste works well across a variety of different model architectures, model sizes and image resolutions. See table \ref{tab:object_size} in the Appendix for benchmark results on different object sizes.}
\label{tab:model_sizes}
\end{table}

\begin{table}[t]
\small
\centering
\begin{tabular}{l|rr}
  \hline
  Setup & Box AP & Mask AP \\
  \hline
  Eff-B7 FPN (640) & 48.5 & 42.7\\
  w/ self-training & \textcolor{blue}{(+1.5)} 50.0 & \textcolor{blue}{(+1.3)} 44.0 \\
  \hline
  w/ Copy-Paste & \textcolor{blue}{(+1.5)} 50.0 & \textcolor{blue}{(+1.0)} 43.7\\
  w/ self-training Copy-Paste & \textcolor{blue}{\textbf{(+2.9)}} \textbf{51.4} & \textcolor{blue}{\textbf{(+2.3)}} \textbf{45.0}\\
  \hline
\end{tabular}
\vspace{1mm}
\caption{Copy-Paste and self-training are additive for utilizing extra unlabeled data.
We get significant improvement of 2.9 box AP and 2.3 mask AP by combining self-training and Copy-Paste.
}
\label{tab:self_sup_results}  
\end{table}

\subsection{Copy-Paste helps data-efficiency}

In this section, we show Copy-Paste is helpful across a variety of dataset sizes and helps data efficiency. Figure~\ref{fig:coco_fraction} reveals that Copy-Paste augmentation is always helpful across all fractions of COCO. Copy-Paste is most helpful in the low data regime (10\% of COCO)  yielding a 6.9 box AP improvement on top of SSJ and a 4.8 box AP improvement on top of LSJ. On the other hand, mixup is only helpful in a low data regime. Copy-Paste also greatly helps with data-efficiency: a model trained on $75\%$ of COCO with Copy-Paste and LSJ has a similar AP to a model trained on $100\%$ of COCO with LSJ.

\subsection{Copy-Paste and self-training are additive}
In this section, we demonstrate that a standard self-training method similar to~\cite{xie2020self, zoph2020rethink} and Copy-Paste can be combined together to leverage unlabeled data. Copy-Paste and self-training individually have similar gains of 1.5 box AP 
over the baseline with 48.5 Box AP (see Table~\ref{tab:self_sup_results}).

To combine self-training and Copy-Paste we first use a supervised teacher model trained with Copy-Paste to generate pseudo labels on unlabeled data. Next we take ground truth objects from COCO and paste them into pseudo labeled images and COCO images. Finally, we train the student model on all these images. With this setup we achieve 51.4 box AP, an improvement of 2.9 AP over the baseline.

\paragraph{Data to Paste on.} In our self-training setup, half of the batch is from supervised COCO data (120k images) and the other half is from pseudo labeled data (110k images from unlabeled COCO and 610k from Objects365). Table~\ref{tab:ablation_selftraining} presents results when we paste COCO instances on different portions of the training images. Pasting into pseudo labeled data yields larger improvements compared to pasting into COCO. Since the number of images in the pseudo labeled set is larger, using images with more variety as background helps Copy-Paste. We get the maximum gain over self-training (+1.4 box AP
) when we paste COCO instances on both COCO and pseudo labeled images.

\paragraph{Data to Copy from.} We also explore an alternative way to use Copy-Paste to incorporate extra data by pasting pseudo labeled objects from an unlabeled dataset directly into the COCO labeled dataset. Unfortunately, this setup shows no additional AP improvements. 

\begin{table}[t]
\small
\centering
\begin{tabular}{lp{1.6cm}|rr}
  \hline
  Setup & Pasting into & Box AP & Mask AP \\
  \hline
  self-training & - & 50.0 &  44.0 \\
  \hline
  +Copy-Paste &  COCO & \textcolor{blue}{(+0.4)} 50.4 & 44.0\\
  +Copy-Paste & Pseudo data & \textcolor{blue}{(+0.8)} 50.8 &  \textcolor{blue}{(+0.5)} 44.5\\
  +Copy-Paste & COCO \& \newline Pseudo data & \textcolor{blue}{(+1.4)} 51.4 &  \textcolor{blue}{(+1.0)} 45.0\\
  \hline
\end{tabular}
\vspace{1mm}
\caption{Pasting ground-truth COCO objects into both COCO and pseudo labeled data gives higher gain in comparison to doing either on its own.}
\label{tab:ablation_selftraining}  
\end{table}


\begin{table*}[t]
\small
\centering
\begin{tabular}{lrrrrrr}
  \hline
  Model &  FLOPs & \# Params & \text{AP$_\text{val}$} & \text{AP$_\text{test-dev}$} & \text{Mask AP$_\text{val}$} & \text{Mask AP$_\text{test-dev}$} \\
  \hline
  SpineNet-190 (1536)~\cite{du2019spinenet} & 2076B & 176M & 52.2  & 52.5 & 46.1 & 46.3\\
  DetectoRS ResNeXt-101-64x4d ~\cite{qiao2020detectors} & --- & --- & --- &55.7$^\dag$  & --- & 48.5 $^\dag$\\
  \hline
  SpineNet-190 (1280)~\cite{du2019spinenet} & 1885B & 164M & 52.6  & 52.8 & --- & ---\\
  SpineNet-190 (1280) w/ self-training~\cite{zoph2019learning} & 1885B & 164M &  54.2 &  54.3 & --- & --- \\
  EfficientDet-D7x (1536)~\cite{tan2019efficientdet} & 410B & 77M & 54.4 & 55.1 & --- & ---\\
  YOLOv4-P7 (1536)~\cite{wang2020scaled}  & --- & --- & --- & 55.8$^\dag$ & --- & ---\\
  \hline
  Cascade Eff-B7 NAS-FPN (1280) & 1440B & 185M  & 54.5  & 54.8 & 46.8 & 46.9\\
  w/ Copy-Paste & 1440B & 185M & \textcolor{blue}{(+1.4)} \textbf{55.9}  & \textcolor{blue}{(+1.2)} \textbf{56.0}  & \textcolor{blue}{(+0.4)} \textbf{47.2} & \textcolor{blue}{(+0.5)} \textbf{47.4}\\
  w/ self-training Copy-Paste & 1440B & 185M & \textcolor{blue}{(+2.5)} \textbf{57.0} & \textcolor{blue}{(+2.5)} \textbf{57.3} & \textcolor{blue}{(+2.1)} \textbf{48.9}&  \textcolor{blue}{(+2.2)} \textbf{49.1}\\
  \hline
\end{tabular}
\vspace{1mm}
\caption{Comparison with the state-of-the-art models on COCO object detection and instance segmentation. Parentheses next to the model name denote the input image size. $^\dag$ indicates results with test time augmentation.}
\label{tab:sota_coco}  
\end{table*}

\subsection{Copy-Paste improves COCO state-of-the-art}

Next we study if Copy-Paste can improve state-of-the-art instance segmentation methods on COCO. Table~\ref{tab:sota_coco} shows the results of applying Copy-Paste on top of a strong 54.8 box AP COCO model. This table is meant to serve as a reference for state-of-the-art performance.\footnote{\url{https://paperswithcode.com/sota/object-detection-on-coco}} For rigorous comparisons, we note that models need to be evaluated with the same codebase, training data, and training settings such as learning rate schedule, weight decay, data pre-processing and augmentations, controlling for parameters and FLOPs, architectural regularization~\cite{wan2013regularization}, training and inference speeds, \etc The goal of the table is to show the benefits of the Copy-Paste augmentation and its additive gains with self-training. Our baseline model is a Cascade Mask-RCNN with EfficientNet-B7 backbone and NAS-FPN. We observe an improvement of +1.2 box AP and +0.5 mask AP using Copy-Paste. When combined with self-training using unlabeled COCO and unlabeled Objects365~\cite{shao2019objects365} for pseudo-labeling, we see a further improvement of 2.5 box AP and 2.2 mask AP, resulting in a model with a strong performance of {\bf 57.3} box AP and {\bf 49.1} mask AP on COCO \texttt{test-dev} without test-time augmentations and model ensembling.

\begin{table}[t]
\small
\centering
\begin{tabular}{lrr}
  \hline
  Model   & AP50 & AP \\
  \hline
  RefineDet512+	\cite{zhang2018single} & 83.8 & -\\
  SNIPER~\cite{singh2018sniper} & 86.9 & -\\
  \hline
  Cascade Eff-B7 NAS-FPN   & 88.6  & 75.0\\
  w/ Copy-Paste pre-training & \textcolor{blue}{(+0.7)} \textbf{89.3}  & \textcolor{blue}{(+1.5)}  \textbf{76.5}\\
  \hline
\end{tabular}
\vspace{1mm}
\caption{PASCAL VOC 2007 detection result on \texttt{test} set. We present results of our EfficientNet-B7 NAS-FPN model pre-trained with and without Copy-Paste on COCO.}
\label{tab:pascal_det_transfer}  
\end{table}

\subsection{Copy-Paste produces better representations for PASCAL detection and segmentation}

Previously we have demonstrated the improved performance that the simple Copy-Paste augmentation provides on instance segmentation. In this section we study the transfer learning performance of the pre-trained instance segmentation models that were trained with Copy-Paste on COCO. Here we perform transfer learning experiments on the PASCAL VOC 2007 dataset. Table~\ref{tab:pascal_det_transfer} shows how the learned Copy-Paste models transfer compared to baseline models on PASCAL detection. Table~\ref{tab:pascal_seg_transfer} shows the transfer learning results on PASCAL semantic segmentation as well. On both PASCAL detection and PASCAL semantic segmentation we find our models trained with Copy-Paste transfer better for fine-tuning than the baseline models.

\begin{table}[t]
\small
\centering
\begin{tabular}{lr}
  \hline
  Model   & mIOU \\
  \hline
  DeepLabv3+ $^\dag$~\cite{chen2018encoder}  & 84.6 \\
  ExFuse $^\dag$~\cite{zhang2018exfuse}  &  85.8 \\
  Eff-B7~\cite{zoph2020rethink} &  85.2 \\
  Eff-L2~\cite{zoph2020rethink} &  88.7 \\
  \hline
  Eff-B7 NAS-FPN & 83.9  \\
  w/ Copy-Paste pre-training & \textcolor{blue}{(+2.7)} \textbf{86.6}  \\
  \hline
\end{tabular}
\vspace{1mm}
\caption{PASCAL VOC 2012 semantic segmentation results on \texttt{val} set. We present results of our EfficientNet-B7 NAS-FPN model pre-trained with and without Copy-Paste on COCO. $^\dag$ indicates multi-scale/flip ensembling inference.
}
\label{tab:pascal_seg_transfer}  
\end{table}

\begin{table*}[t]
\centering
\small
\begin{tabular}{l|rrrr|rrrr}
 \hline
 & Mask AP & Mask AP$_\text{r}$ & Mask AP$_\text{c}$ & Mask AP$_\text{f}$ & Box AP \\
 \hline
 cRT (ResNeXt-101-32$\times$8d)~\cite{kang2020decoupling} & 27.2 & 19.6 & 26.0 & 31.9 & --- \\ 
 LVIS Challenge 2020 Winner$^\dagger$~\cite{lviswinner} & 38.8 & 28.5 & 39.5 & 42.7 & 41.1 \\ 
 \hline
 ResNet-50 FPN (1024) & 30.3 & 22.2 & 29.5 & 34.7 & 31.5 \\
 w/ Copy-Paste & \textcolor{blue}{(+2.0)} 32.3 & \textcolor{blue}{(+4.3)} 26.5 & \textcolor{blue}{(+2.3)} 31.8 & \textcolor{blue}{(+0.6)} 35.3 & \textcolor{blue}{(+2.8)} 34.3 \\ 
 \hline
 ResNet-101 FPN (1024) & 31.9 & 24.7 & 30.5 & 36.3 & 33.3 \\ 
 w/ Copy-Paste & \textcolor{blue}{(+2.1)} 34.0 & \textcolor{blue}{(+2.7)} 27.4 & \textcolor{blue}{(+3.4)} 33.9 & \textcolor{blue}{(+0.9)} 37.2 & \textcolor{blue}{(+3.1)} 36.4 \\ 
 \hline
 EfficientNet-B7 FPN (1024) & 33.7 & 26.4 & 33.1 & 37.6 & 35.5 \\ 
 w/ Copy-Paste & \textcolor{blue}{(+2.3)} 36.0 & \textcolor{blue}{(+3.3)} 29.7 & \textcolor{blue}{(+2.7)} 35.8 & \textcolor{blue}{(+1.3)} 38.9 & \textcolor{blue}{(+3.7)} 39.2 \\ 
 \hline
 EfficientNet-B7 NAS-FPN (1280) & 34.7 & 26.0 & 33.4 & 39.8 & 37.2 \\  
 w/ Copy-Paste  & \textcolor{blue}{(+3.4)} 38.1 & \textcolor{blue}{(+6.1)} 32.1 & \textcolor{blue}{(+3.7)} 37.1 & \textcolor{blue}{(+2.1)} 41.9 & \textcolor{blue}{(+4.4)} 41.6 \\ 
 \hline
\end{tabular}
\vspace{1mm}
\caption{Comparison with the state-of-the-art models on LVIS v1.0 object detection and instance segmentation. Parentheses next to our models denote the input image size. $^\dagger$ We report the 2020 winning entry's result without test-time augmentation.}
\label{tab:sota_lvis}  
\end{table*}

\subsection{Copy-Paste provides strong gains on LVIS}

We benchmark Copy-Paste on the LVIS dataset to see how it performs on a dataset with a long-tail distribution of 1203 classes. There are two different training paradigms typically used for LVIS: (1) single-stage where a detector is trained directly on the LVIS dataset, (2) two-stage where the model from the first stage is fine-tuned with class re-balancing losses to help handle the class imbalance.

\paragraph{Copy-Paste improves single-stage LVIS training.}

The single-stage training paradigm is quite similar to our Copy-Paste setup on COCO. In addition to the standard training setup, certain methods are used to handle the class imbalance problem on LVIS. One common method is Repeat Factor Sampling (RFS) from~\cite{gupta2019lvis}, with $t=0.001$. This method aims at helping the large class imbalance problem on LVIS by over-sampling images that contain less frequent object categories. 
For single-stage training on LVIS, we follow the same training parameters on COCO to train our models for 180k steps using a 256 batch size.
As suggested by~\cite{gupta2019lvis}, we increase the number of detections per image to 300 and reduce the score threshold to 0.
Table~\ref{tab:lvis-single-stage} shows the results of applying Copy-Paste to a strong single-stage LVIS baseline of EfficientNet-B7 FPN with 640$\times$640 input size. 
We observe that Copy-Paste augmentation outperforms RFS on AP, AP$_\text{c}$ and AP$_\text{f}$, but under-performs on AP$_\text{r}$ (the AP for rare classes). The best overall result comes from combining RFS and Copy-Paste augmentation, achieving a boost of +2.4 AP and +8.7 AP$_\text{r}$.

\begin{table}[t]
\small
\centering
\begin{tabular}{l|r|rrr}
  \hline
  Setup (single-stage) & AP & AP$_\text{r}$ & AP$_\text{c}$ & AP$_\text{f}$ \\
  \hline
  Eff-B7 FPN (640) & 27.7 & 9.7 & 28.1 & 35.1 \\
  w/ RFS & 28.2 & 15.4 & 27.8 & 34.3 \\
  w/ Copy-Paste & 29.3 & 12.8 & \textbf{30.1} & \textbf{35.7} \\
  w/ RFS w/ Copy-Paste & \textbf{30.1} & \textbf{18.4} & 30.0 & 35.4 \\
  \hline
\end{tabular}
\vspace{1mm}
\caption{Single-stage training results (mask AP) on LVIS. 
}
\label{tab:lvis-single-stage}
\end{table}

\paragraph{Copy-Paste improves two-stage LVIS training.}

Two-stage training is widely adopted to address data imbalance and obtain good performance on LVIS~\cite{li2020overcoming,ren2020balanced,lviswinner}. We aim to study the efficacy of Copy-Paste in this two-stage setup. Our two-stage training is as follows: first we train the object detector with standard training techniques (\ie, same as our single-stage training) and then we fine-tune the model trained in the first stage using the Class-Balanced Loss~\cite{cui2019class}. The weight for a class is calculated by $(1-\beta)/(1-\beta^n)$, where $n$ is the number of instances of the class and $\beta = 0.999$.\footnote{We scale class weights by dividing the mean and then clip their values to $[0.01, 5]$, as suggested by~\cite{li2020overcoming}.}
During the second stage fine-tuning, we train the model with 3$\times$ schedule and only update the final classification layer in Mask R-CNN using the classification loss only.
From mask AP results in Table~\ref{tab:lvis-two-stage}, we can see models trained with Copy-Paste learn better features for low-shot classes (+2.3 on AP$_\text{r}$ and +2.6 on AP$_\text{c}$).
Interestingly, we find RFS, which is quite helpful and additive with Copy-Paste in single-stage training, hurts the performance in two-stage training. A possible explanation for this finding is that features learned with RFS are worse than those learned with the original LVIS dataset. We leave a more detailed investigation of the tradeoffs between RFS and data augmentations in two stage training for future work.

\begin{table}[t]
\small
\centering
\begin{tabular}{l|r|rrr} 
  \hline
  Setup (two-stage) & AP & AP$_\text{r}$ & AP$_\text{c}$ & AP$_\text{f}$ \\
  \hline
  Eff-B7 FPN (640) & 31.3 & 25.0 & 30.6 & 34.9 \\
  w/ RFS & 30.1 & 21.8 & 29.7 & 34.1 \\
  w/ Copy-Paste & \textbf{33.0} & \textbf{27.3} & \textbf{33.2} & \textbf{35.7} \\
  w/ RFS w/ Copy-Paste & 32.0 & 26.3 & 31.8 & 34.7 \\
  \hline
\end{tabular}
\vspace{1mm}
\caption{Two-stage training results (mask AP) on LVIS.}
\label{tab:lvis-two-stage}
\end{table}

\paragraph{Comparison with the state-of-the-art.} Furthermore, we compare our two-stage models with state-of-the-art methods for LVIS\footnote{\url{https://www.lvisdataset.org/challenge_2020}} in Table~\ref{tab:sota_lvis}.
Surprisingly, our smallest model, ResNet-50 FPN, outperforms a strong baseline cRT~\cite{kang2020decoupling} with ResNeXt-101-32$\times$8d backbone.

EfficientNet-B7 NAS-FPN model (without Cascade~\footnote{We find using Cascade in our experiments improves AP$_\text{f}$ but hurts AP$_\text{r}$.}) trained with Copy-Paste achieves comparable performance to LVIS challenge 2020 winner on overall Mask AP and Box AP without test-time augmentation. Also, it obtains 32.1 mask AP$_\text{r}$ for rare categories, outperforming the LVIS Challenge 2020 winning entry by +3.6 mask AP$_\text{r}$.

\section{Conclusion}

Data augmentation is at the heart of many vision systems. In this paper, we rigorously studied the Copy-Paste data augmentation method, and found that it is very effective and robust. Copy-Paste performs well across multiple experimental settings and provides significant improvements on top of strong baselines, both on the COCO and LVIS instance segmentation benchmarks. 

The Copy-Paste augmentation strategy is simple, easy to plug into any instance segmentation codebase, and does not increase the training cost or inference time. We also showed that Copy-Paste is useful for incorporating extra unlabeled images during training and is additive on top of successful self-training techniques. We hope that the convincing empirical evidence of its benefits make Copy-Paste augmentation a standard augmentation procedure when training instance segmentation models.

{\small
\bibliographystyle{ieee_fullname}
\bibliography{main}
}

\clearpage
\include{appendix}

\end{document}

%% file: abstract.tex
\begin{abstract}

Building instance segmentation models that are data-efficient and can handle rare object categories is an important challenge in computer vision. Leveraging data augmentations is a promising direction towards addressing this challenge. Here, we perform a systematic study of the Copy-Paste augmentation (\eg,~\cite{dwibedi2017cut, dvornik2018modeling}) for instance segmentation where we randomly paste objects onto an image. Prior studies on Copy-Paste relied on modeling the surrounding visual context for pasting the objects. However, we find that the simple mechanism of pasting objects randomly is good enough and can provide solid gains on top of strong baselines. Furthermore, we show Copy-Paste is additive with semi-supervised methods that leverage extra data through pseudo labeling (\eg self-training). On COCO instance segmentation, we achieve 49.1 mask AP and 57.3 box AP, an improvement of +0.6 mask AP and +1.5 box AP over the previous state-of-the-art. We further demonstrate that Copy-Paste can lead to significant improvements on the LVIS benchmark. Our baseline model outperforms the LVIS 2020 Challenge winning entry by +3.6 mask AP on rare categories. \footnote{Code and checkpoints for our models are available at \color{magenta}\url{https://github.com/tensorflow/tpu/tree/master/models/official/detection/projects/copy_paste}} \looseness=-1

\end{abstract}

%% file: introduction.tex
\section{Introduction}

\begin{figure}[t]
    \centering
    \includegraphics[width=0.9\linewidth]{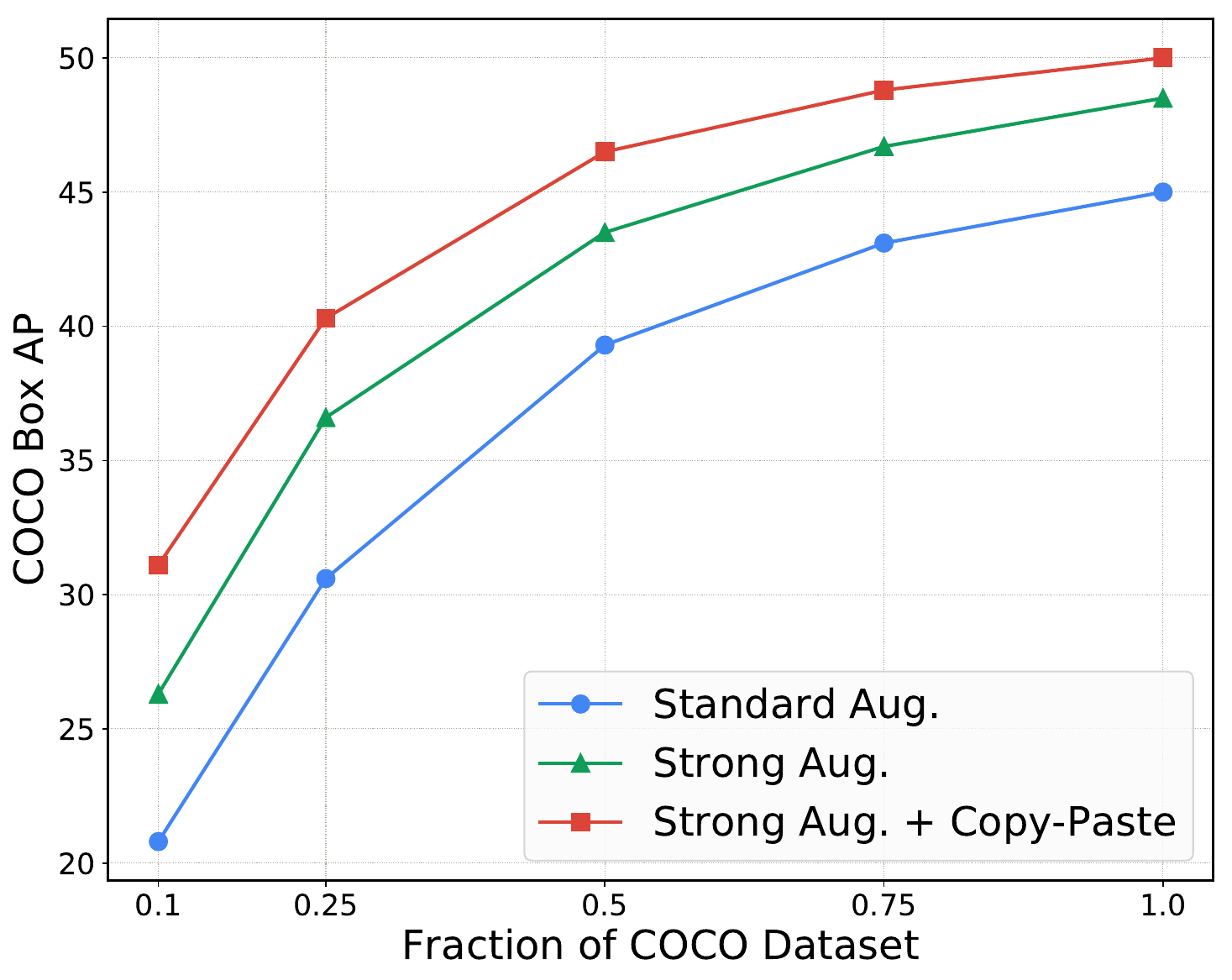}
    \caption{Data-efficiency on the COCO benchmark: Combining the Copy-Paste augmentation along with Strong Aug. (large scale jittering) allows us to train models that are up to 2$\times$ more data-efficient than Standard Aug. (standard scale jittering). The augmentations are highly effective and provide gains of +10 AP in the low data regime (10\% of data) while still being effective in the high data regime with a gain of +5 AP. Results are for Mask R-CNN EfficientNet-B7 FPN trained on an image size of 640$\times$640.
    }
    \label{fig:teaser}
\vspace{-2mm}
\end{figure}

Instance segmentation~\cite{hariharan2014simultaneous, dai2016instance} is an important task in computer vision with many real world applications.  
Instance segmentation models based on state-of-the-art convolutional networks~\cite{du2019spinenet, tan2019efficientdet, zhang2020resnest} are often data-hungry. At the same time, annotating large datasets for instance segmentation~\cite{lin2014microsoft,gupta2019lvis} is usually expensive and time-consuming. For example, 22 worker hours were spent per 1000 instance masks for COCO~\cite{lin2014microsoft}. It is therefore imperative to develop new methods to improve the data-efficiency of state-of-the-art instance segmentation models.  

\begin{figure*}[t]
    \centering
    \includegraphics[width=0.75\textwidth]{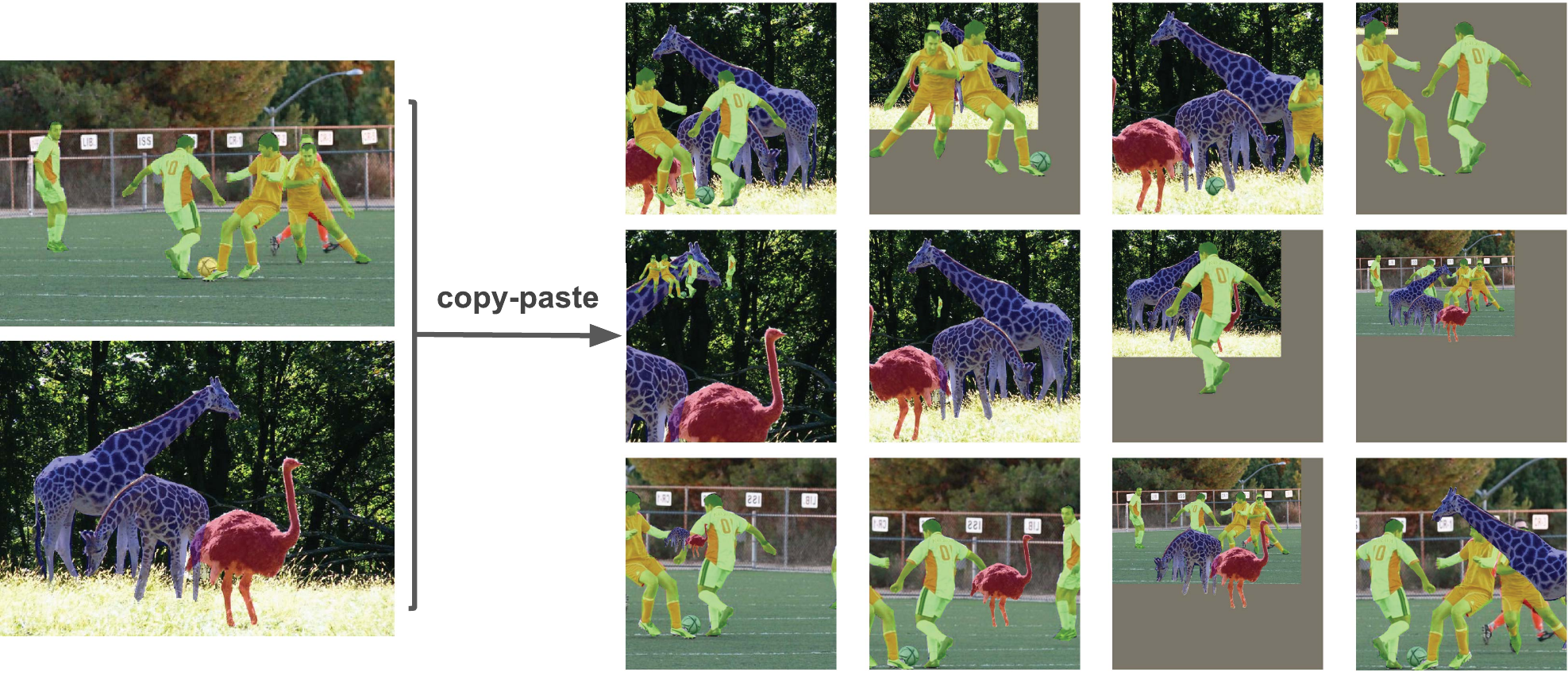}
    \caption{We use a simple copy and paste method to create new images for training instance segmentation models. We apply random scale jittering on two random training images and then randomly select a subset of instances from one image to paste onto the other image.}
    \label{fig:method}
\vspace{-2mm}
\end{figure*}

Here, we focus on data augmentation~\cite{shorten2019survey} as a simple way to significantly improve the data-efficiency of instance segmentation models. 
Although many augmentation methods such as scale jittering and random resizing have been widely used~\cite{he2017mask, he2019rethinking,girshick2018detectron}, they are more general-purpose in nature and have not been designed specifically for instance segmentation. An augmentation procedure that is more \emph{object-aware}, both in terms of category and shape, is likely to be useful for instance segmentation. The Copy-Paste augmentation~\cite{dwibedi2017cut, dvornik2018modeling, fang2019instaboost}  is well suited for this need. By pasting diverse objects of various scales to new background images, Copy-Paste has the potential to create challenging and novel training data for free.

The key idea behind the Copy-Paste augmentation is to paste objects from one image to another image. This can lead to a {\it combinatorial} number of new training data, with multiple possibilities for: (1) choices of the pair of source image from which instances are copied, and the target image on which they are pasted; (2) choices of object instances to copy from the source image; (3) choices of where to paste the copied instances on the target image. 
The large variety of options when utilizing this data augmentation method allows for lots of exploration on how to use the technique most effectively.
Prior work~\cite{dvornik2018modeling, fang2019instaboost} adopts methods for deciding where to paste the additional objects by modeling the surrounding visual context. In contrast, we find that a simple strategy of randomly picking objects and pasting them at random locations on the target image provides a significant boost on top of baselines across multiple settings. Specifically, it gives solid improvements across a wide range of settings with variability in backbone architecture, extent of scale jittering, training schedule and image size.

In combination with large scale jittering, we show that the Copy-Paste augmentation results in significant gains in the data-efficiency on COCO (Figure~\ref{fig:teaser}). In particular, we see a data-efficiency improvement of 2$\times$ over the commonly used standard scale jittering data augmentation. We also observe a gain of +10 Box AP on the low-data regime when using only 10\% of the COCO training data.

We then show that the Copy-Paste augmentation strategy provides additional gains with self-training~\cite{radosavovic2018data,zoph2020rethink} wherein we extract instances from ground-truth data and paste them onto unlabeled data annotated with pseudo-labels.
Using an EfficientNet-B7~\cite{tan2019efficientnet} backbone and NAS-FPN~\cite{ghiasi2019fpn} architecture, we achieve 57.3 Box AP and 49.1 Mask AP on COCO \texttt{test-dev} without test-time augmentations. This result surpasses the previous state-of-the-art instance segmentation models such as SpineNet~\cite{du2019spinenet} (46.3 mask AP) and DetectoRS ResNeXt-101-64x4d with test time augmentation~\cite{qiao2020detectors} (48.5 mask AP). The performance also surpasses state-of-the-art bounding box detection results of EfficientDet-D7x-1536~\cite{tan2019efficientdet} (55.1 box AP) and YOLOv4-P7-1536~\cite{wang2020scaled} (55.8 box AP) despite using a smaller image size of 1280 instead of 1536.

Finally, we show that the Copy-Paste augmentation results in better features for the two-stage training procedure typically used in the LVIS benchmark~\cite{gupta2019lvis}. Using Copy-Paste we get improvements of 6.1 and 3.7 mask AP on the rare and common categories, respectively.

The Copy-Paste augmentation strategy is easy to plug into any instance segmentation codebase, can utilize unlabeled images effectively and does not create training or inference overheads. For example, our experiments with Mask-RCNN show that we can drop Copy-Paste into its training, and without any changes, the results can be easily improved, \eg, by +1.0 AP for 48 epochs.

%% file: related.tex
\section{Related Work}

\paragraph{Data Augmentations.}
Compared to the volume of work on backbone architectures~\cite{krizhevsky2012, simonyan2014very, szegedy2015, he2016deep, tan2019efficientnet} and detection/segmentation frameworks~\cite{girshick2014rich, girshick2015fast, ren2015faster, lin2017feature, he2017mask, lin2017focal}, relatively less attention is paid to data augmentations~\cite{shorten2019survey} in the computer vision community. Data augmentations such as random crop~\cite{lecun98gradient, krizhevsky2012, simonyan2014very, szegedy2015}, color jittering~\cite{szegedy2015}, Auto/RandAugment~\cite{cubuk2018autoaugment, cubuk2020randaugment} have played a big role in achieving state-of-the-art results on image classification~\cite{he2016deep, tan2019efficientnet}, self-supervised learning~\cite{henaff2019data, he2020momentum, chen2020simple} and semi-supervised learning~\cite{xie2020self} on the ImageNet~\cite{russakovsky2015imagenet} benchmark. These augmentations are more general purpose in nature and are mainly used for encoding {\it invariances to data transformations}, a principle well suited for image classification~\cite{russakovsky2015imagenet}.

\paragraph{Mixing Image Augmentations.}
In contrast to augmentations that encode invariances to data transformations, there exists a class of augmentations that mix the information contained in different images with appropriate changes to groundtruth labels. A classic example is the mixup data augmentation~\cite{zhang2017mixup} method which creates new data points for free from convex combinations of the input pixels and the output labels. There have been adaptations of mixup such as CutMix~\cite{yun2019cutmix} that pastes rectangular crops of an image instead of mixing all pixels. There have also been applications of mixup and CutMix to object detection~\cite{zhang2019bag}. The Mosaic data augmentation method employed in YOLO-v4~\cite{bochkovskiy2020yolov4} is related to CutMix in the sense that one creates a new compound image that is a rectangular grid of multiple individual images along with their ground truths. While mixup, CutMix and Mosaic are useful in combining multiple images or their cropped versions to create new training data, they are still not {\it object-aware} and have not been designed specifically for the task of instance segmentation.

\paragraph{Copy-Paste Augmentation.}
A simple way to combine information from multiple images in an {\it object-aware} manner is to copy instances of objects from one image and paste them onto another image. Copy-Paste is akin to mixup and CutMix but only copying the exact pixels corresponding to an object as opposed to all pixels in the object's bounding box. 
One key difference in our work compared to Contextual Copy-Paste~\cite{dvornik2018modeling} and InstaBoost~\cite{fang2019instaboost} is that we do not need to model surrounding visual context to place the copied object instances. A simple random placement strategy works well and yields solid improvements on strong baseline models. Instaboost~\cite{fang2019instaboost} differs from prior work on Copy-Paste~\cite{dvornik2018modeling} by not pasting instances from other images but rather by jiterring instances that already exist on the image. Cut-Paste-and-Learn~\cite{dwibedi2017cut} proposes to extract object instances, blend and paste them on diverse backgrounds and train on the augmented images in addition to the original dataset. Our work uses the same method with some differences: (1) We do not use geometric transformations (\eg rotation), and find Gaussian blurring of the pasted instances not beneficial; (2) We study Copy-Paste in the context of pasting objects contained in one image into another image already populated with instances where~\cite{dwibedi2017cut} studies Copy-Paste in the context of having a bank of object instances and background scenes to improve performance;
(3) We study the efficacy of Copy-Paste in the semi-supervised learning setting by using it in conjunction with self-training. 
(4) We benchmark and thoroughly study Copy-Paste on the widely used COCO and LVIS datasets while Cut-Paste-and-Learn uses the GMU dataset~\cite{georgakis2016multiview}. A key contribution is that our paper shows the use of Copy-Paste in improving state-of-the-art instance segmentation models on COCO and LVIS.

\paragraph{Instance Segmentation.}
Instance segmentation~\cite{hariharan2014simultaneous, hariharan2015hypercolumns} is a challenging computer vision problem that attempts to both detect object instances and segment the pixels corresponding to each instance. Mask-RCNN~\cite{he2017mask} is a widely used framework with most state-of-the-art methods~\cite{zhang2020resnest, du2019spinenet, qiao2020detectors} adopting that approach. The COCO dataset is the widely used benchmark for measuring progress. We report state-of-the-art\footnote{Based on the entries in \url{https://paperswithcode.com/sota/instance-segmentation-on-coco}.}  results on the COCO benchmark surpassing SpineNet~\cite{du2019spinenet} by 2.8 AP and DetectoRS~\cite{qiao2020detectors} by 0.6 AP.\footnote{We note that better mask / box AP on COCO have been reported in COCO competitions in 2019 - \url{https://cocodataset.org/workshop/coco-mapillary-iccv-2019.html}.} 

Copy and paste approach is also used for weakly supervised instance segmentation. Remez et al. \cite{remez2018learning} introduce an adversarial approach where it uses a generator network to predict the segmentation mask of an object within a given bounding box. Given the generated mask, the object is blended on another background and then a discriminator network is used to make sure the generated mask/image looks realistic. Different from this work, we use Copy-Paste as an augmentation method.

\paragraph{Long-Tail Visual Recognition.}
Recently, the computer vision community has begun to focus on the long-tail nature of object categories present in natural images~\cite{van2018inaturalist,gupta2019lvis}, where many of the different object categories have very few labeled images.
Modern approaches for addressing long-tail data when training deep networks can be mainly divided into two groups: data re-sampling~\cite{mahajan2018exploring,gupta2019lvis,wang2020devil} and loss re-weighting~\cite{huang2016learning,cui2019class,cao2019learning,tan2020equalization,li2020overcoming,ren2020balanced}.
Other more complicated learning methods (\eg, meta-learning~\cite{wang2019meta,hu2020learning,jamal2020rethinking}, causal inference~\cite{tang2020long}, Bayesian methods~\cite{khan2019striking}, \etc) are also used to deal with long-tail data.
Recent work~\cite{cui2018large,cao2019learning,kang2020decoupling,zhou2020bbn,li2020overcoming} has pointed out the effectiveness of two-stage training strategies by separating the feature learning and the re-balancing stage, as end-to-end training with re-balancing strategies could be detrimental to feature learning.
A more comprehensive summary of data imbalance in object detection can be found in Oksuz~\etal~\cite{oksuz2020imbalance}.
Our work demonstrates simple Copy-Paste data augmentation yields significant gains in both single-stage and two-stage training on the LVIS benchmark, especially for rare object categories.

%% file: appendix.tex
\begin{appendix}

\section{Ablation on the Copy-Paste method}
In this section we present ablations for our Copy-Paste method. We use Mask R-CNN EfficientNetB7-FPN architecture and image size of 640$\times$640 for our experiments.

\paragraph{Subset of pasted objects.} In our method, we paste a \textit{random} subset of objects from one image onto another image. Table~\ref{tab:ablation} shows that although we get improvements from pasting only \textit{one} random object or \textit{all} the objects of one image into another image, we get the best improvement by pasting a \textit{random} subset of objects. This shows that the added randomness introduced from pasting a subset of objects is helpful.

\paragraph{Blending.} In our experiments, we smooth out the edges of pasted objects using alpha blending (see Section~\ref{sec:method}). Table~\ref{tab:ablation} shows that this is not an important step and we get the same results without any blending in contrast to~\cite{dwibedi2017cut} who find blending is crucial for strong performance.

\begin{table}[h!]
\centering
\begin{tabular}{l|rr}
  \hline
  Setup & Box AP & Mask AP \\
  \hline
  EfficientNetB7-FPN (640) & 48.5 & 42.7\\
  \hline
  w/ Copy-Paste (one object) & \textcolor{red}{(-0.9)} 49.1  & \textcolor{red}{(-0.6)} 43.1\\
  w/ Copy-Paste (all objects) & \textcolor{red}{(-0.3)} 49.7 & \textcolor{red}{(-0.4)} 43.3 \\
  \hline
  w/ Copy-Paste (no blending) & \textbf{50.0} &  \textbf{43.7}\\
  w/ Copy-Paste & \textbf{50.0} & \textbf{43.7}\\
  \hline
\end{tabular}
\vspace{1mm}
\caption{Ablation studies for the Copy-Paste method on COCO. We study the value of applying blending to pasted objects along with how many objects to paste from one image to another.}
\label{tab:ablation}  
\end{table}

\paragraph{Scale jittering.} In this work, we show that by combining large scale jittering and Copy-Paste we obtain a significant improvement over the baseline with standard scale jittering (Figure~\ref{fig:teaser}). In the Copy-Paste method, we apply independent random scale jittering on both the pasted image (image that pasted objects are being copied from) and the main image. In Table~\ref{tab:ablation_sj} we study the importance of large scale jittering on both the main and the pasted images. Table~\ref{tab:ablation_sj} shows that most of the improvement from large scale jittering is coming from applying it on the main image and we only get slight improvement (0.3 box AP and 0.2 Mask AP) from increasing the scale jittering range for the pasted image.

\begin{table}[h!]
\centering
\begin{tabular}{ll|rr}
  \hline
  Main Image & Pasted Image & Box AP & Mask AP \\
  \hline

  SSJ& SSJ  & \textcolor{red}{(-1.9)} 48.1 & \textcolor{red}{(-1.6)} 42.1\\
  SSJ& LSJ  & \textcolor{red}{(-2.3)} 47.7  & \textcolor{red}{(-1.9)} 41.8 \\
  LSJ& SSJ  & \textcolor{red}{(-0.3)} 49.7 & \textcolor{red}{(-0.2)} 43.5 \\
  LSJ & LSJ  & \textbf{50.0} & \textbf{43.7}\\
  \hline
\end{tabular}
\vspace{1mm}
\caption{Ablation study on scale jittering methods for the main image and the pasted image.}
\label{tab:ablation_sj}  
\end{table}

\section{Copy-Paste provides more gain on harder categories of COCO}
\begin{figure*}[t!]
  \centering
    \begin{tabular}{c}
    \includegraphics[width=1.0\linewidth,origin=c]{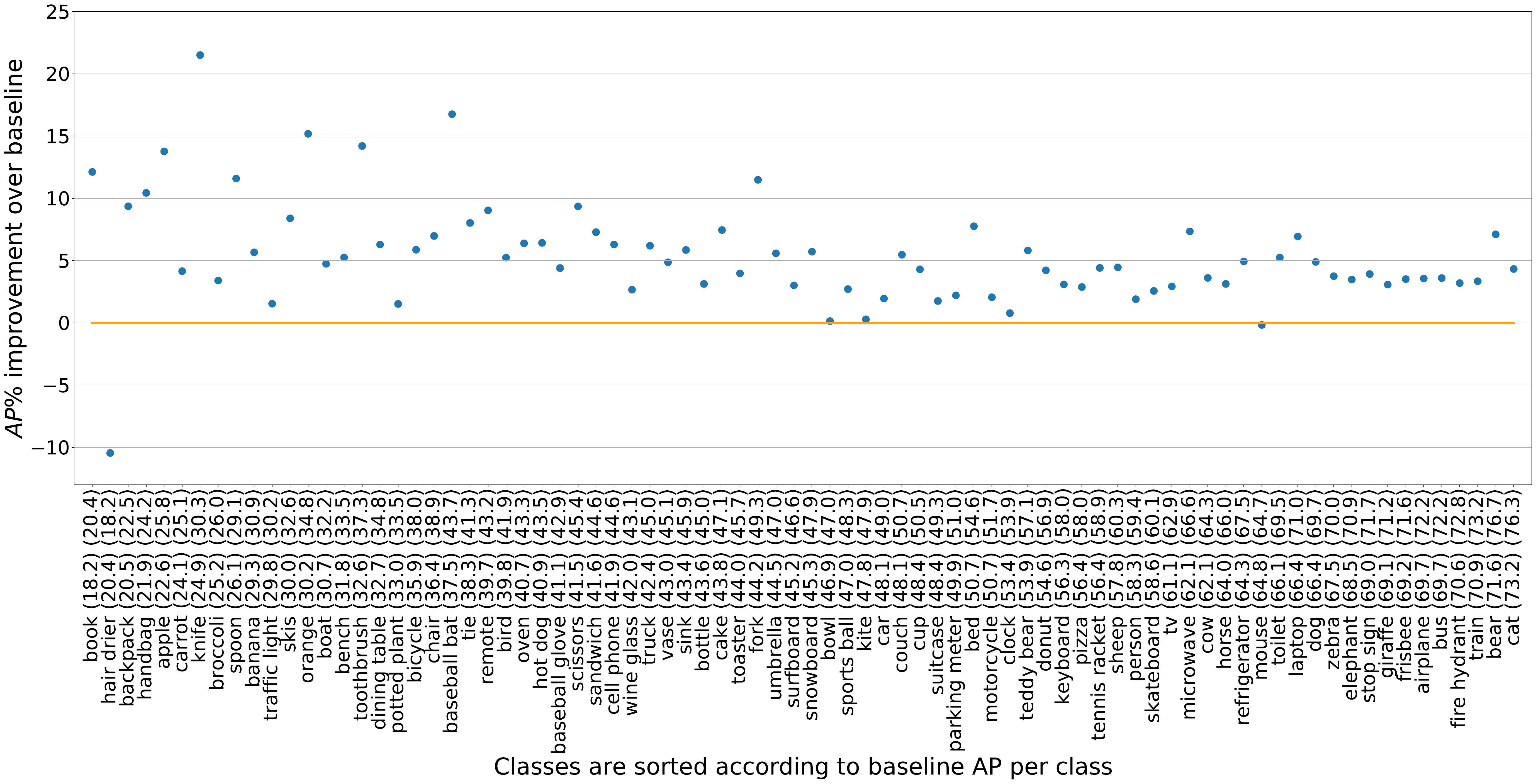}\\
    \end{tabular}
\caption{Per category relative AP improvement from Copy-Paste on 80 classes of COCO dataset. Numbers in the parentheses show the AP per category of the baseline model (first number) and the model trained with Copy-Paste (second number). Each number is the average over 5 runs.
Classes are sorted based on the baseline AP per class.
}
\label{fig:per_category}
\end{figure*}

Figure~\ref{fig:per_category} shows the relative AP gain per category obtained from applying Copy-Paste on the COCO dataset. Copy-Paste improves the AP of all the classes except hair drier. In Figure~\ref{fig:per_category} classes are sorted based on the baseline AP per category. We observe most of the classes with the highest improvement are on the left (lower baseline AP) which shows Copy-Paste helps the hardest classes the most.


\section{How likely objects are copied to an unmatched scene?}
In our method, we copy objects from a random image to another random image without considering the context of the images. In this section we compute the probability of copying objects to an unmatched scene category (context) of indoor or outdoor.

COCO images do not have scene categories. But, we use COCO-panoptic labels to assign the COCO images to indoor or outdoor scene categories. We found there are 42538 indoor and 71017 outdoor images (we couldn't estimate the category of the rest 4732 images).
Table~\ref{tab:scene} shows the probability of copying objects from one scene category to another. Therefore, we copy objects to an unmatched scene in about half (46.8\%) of generated images. 

\begin{table}[h]
\centering
\begin{tabular}{c|cc}
\diagbox{from}{to} & indoor & outdoor \\
   \hline
   indoor  &  14.1\%  & 23.4\%\\
   outdoor &  23.4\%  & 39.1\%
\end{tabular}
\caption{Probability of copying objects from one scene category to another scene category for COCO dataset.}
\label{tab:scene}
\vspace{-2mm}
\end{table}

\section{Benchmark results on different object sizes}
In the table \ref{tab:model_sizes} we report Copy-paste performance of variety of model architectures. In table \ref{tab:object_size} we provide additional benchmarks on different object sizes.
\label{sec:object_sizes}

\begin{table*}[t]
\small
\centering
\begin{tabular}{lrrrrrrrr}
  \hline
  Model  &  box AP & \text{box AP$_\text{s}$} & \text{box AP$_\text{m}$} & \text{box AP$_\text{l}$} & mask AP & \text{mask AP$_\text{s}$} & \text{mask AP$_\text{m}$} & \text{mask AP$_\text{l}$}\\
  \hline
  Res-50 FPN (1024) & 47.2 & 28.5 & 49.6 & 64.6 & 41.8 & 23.0 & 44.3 & 60.1\\
  w/ Copy-Paste  & \textcolor{blue}{(+1.0)} 48.2& \textcolor{blue}{(+0.4)} 28.9 & \textcolor{blue}{(+1.4)} 51.0 & \textcolor{blue}{(+1.8)} 66.4 & \textcolor{blue}{(+0.6)}  42.4 & \textcolor{blue}{(+0.2)} 23.2 & \textcolor{blue}{(+0.9)} 45.2 & \textcolor{blue}{(+1.1)} 61.2\\
  \hline
  Res-101 FPN (1024) & 48.4   & 29.2 & 51.1 & 65.8 & 42.8 & 23.5 & 45.5 & 60.4\\
  w/ Copy-Paste  & \textcolor{blue}{(+1.4)} 49.8  & \textcolor{blue}{(+1.3)} 30.5  & \textcolor{blue}{(+1.8)} 52.9  & \textcolor{blue}{(+1.1)} 66.9 & \textcolor{blue}{(+0.8)} 43.6 & \textcolor{blue}{(+1.0)} 24.5 & \textcolor{blue}{(+1.4)} 46.9 & \textcolor{blue}{(+1.1)} 61.5\\
  \hline
  Res-101 FPN (1280)  & 49.1 & 30.4 & 51.9 & 66.6 & 43.1 & 24.5 & 46.0 & 61.6\\
  w/ Copy-Paste &  \textcolor{blue}{(+1.2)} 50.3 & \textcolor{blue}{(+1.3)} 31.7 & \textcolor{blue}{(+1.8)} 53.7 & \textcolor{blue}{(+0.6)} 67.2 & \textcolor{blue}{(+1.1)} 44.2 & \textcolor{blue}{(+1.2)} 25.7 & \textcolor{blue}{(+1.5)} 47.5 & \textcolor{blue}{(+0.2)} 61.8\\
  \hline
  Eff-B7 FPN  (1280)  & 51.1 & 33.3 & 53.9 & 67.9 & 44.8 & 26.6 & 47.9 & 62.7 \\
  w/ Copy-Paste  & \textcolor{blue}{(+1.5)} 52.6 & \textcolor{blue}{(+1.0)} 34.3 & \textcolor{blue}{(+1.7)} 55.6 & \textcolor{blue}{(+2.3)} 70.2 & \textcolor{blue}{(+1.1)} 45.9 & \textcolor{blue}{(+0.9)} 27.5 & \textcolor{blue}{(+1.5)} 49.4 & \textcolor{blue}{(+1.8)} 64.5\\
  \hline
\end{tabular}
\vspace{1mm}
\caption{Box AP and Mask AP benchmark results on different object sizes for models trained with different backbones and image sizes.}
\label{tab:object_size}
\end{table*}

\end{appendix}

%% file: main.bbl
\begin{thebibliography}{10}\itemsep=-1pt

\bibitem{bochkovskiy2020yolov4}
Alexey Bochkovskiy, Chien-Yao Wang, and Hong-Yuan~Mark Liao.
\newblock Yolov4: Optimal speed and accuracy of object detection.
\newblock {\em arXiv preprint arXiv:2004.10934}, 2020.

\bibitem{cai2018cascade}
Zhaowei Cai and Nuno Vasconcelos.
\newblock Cascade r-cnn: Delving into high quality object detection.
\newblock In {\em CVPR}, 2018.

\bibitem{cao2019learning}
Kaidi Cao, Colin Wei, Adrien Gaidon, Nikos Arechiga, and Tengyu Ma.
\newblock Learning imbalanced datasets with label-distribution-aware margin
  loss.
\newblock In {\em NeurIPS}, 2019.

\bibitem{chen2018encoder}
Liang-Chieh Chen, Yukun Zhu, George Papandreou, Florian Schroff, and Hartwig
  Adam.
\newblock Encoder-decoder with atrous separable convolution for semantic image
  segmentation.
\newblock In {\em ECCV}, 2018.

\bibitem{chen2020simple}
Ting Chen, Simon Kornblith, Mohammad Norouzi, and Geoffrey Hinton.
\newblock A simple framework for contrastive learning of visual
  representations.
\newblock In {\em ICML}, 2020.

\bibitem{cubuk2018autoaugment}
Ekin~D Cubuk, Barret Zoph, Dandelion Mane, Vijay Vasudevan, and Quoc~V Le.
\newblock Autoaugment: Learning augmentation policies from data.
\newblock In {\em CVPR}, 2019.

\bibitem{cubuk2020randaugment}
Ekin~D Cubuk, Barret Zoph, Jonathon Shlens, and Quoc~V Le.
\newblock Randaugment: Practical automated data augmentation with a reduced
  search space.
\newblock In {\em NeurIPS}, 2020.

\bibitem{cui2019class}
Yin Cui, Menglin Jia, Tsung-Yi Lin, Yang Song, and Serge Belongie.
\newblock Class-balanced loss based on effective number of samples.
\newblock In {\em CVPR}, 2019.

\bibitem{cui2018large}
Yin Cui, Yang Song, Chen Sun, Andrew Howard, and Serge Belongie.
\newblock Large scale fine-grained categorization and domain-specific transfer
  learning.
\newblock In {\em CVPR}, 2018.

\bibitem{dai2016instance}
Jifeng Dai, Kaiming He, and Jian Sun.
\newblock Instance-aware semantic segmentation via multi-task network cascades.
\newblock In {\em CVPR}, 2016.

\bibitem{du2019spinenet}
Xianzhi Du, Tsung-Yi Lin, Pengchong Jin, Golnaz Ghiasi, Mingxing Tan, Yin Cui,
  Quoc~V Le, and Xiaodan Song.
\newblock Spinenet: Learning scale-permuted backbone for recognition and
  localization.
\newblock In {\em CVPR}, 2020.

\bibitem{dvornik2018modeling}
Nikita Dvornik, Julien Mairal, and Cordelia Schmid.
\newblock Modeling visual context is key to augmenting object detection
  datasets.
\newblock In {\em ECCV}, 2018.

\bibitem{dwibedi2017cut}
Debidatta Dwibedi, Ishan Misra, and Martial Hebert.
\newblock Cut, paste and learn: Surprisingly easy synthesis for instance
  detection.
\newblock In {\em ICCV}, 2017.

\bibitem{pascal}
Mark Everingham, Luc Van~Gool, Christopher~KI Williams, John Winn, and Andrew
  Zisserman.
\newblock The pascal visual object classes (voc) challenge.
\newblock {\em IJCV}, 2010.

\bibitem{fang2019instaboost}
Hao-Shu Fang, Jianhua Sun, Runzhong Wang, Minghao Gou, Yong-Lu Li, and Cewu Lu.
\newblock Instaboost: Boosting instance segmentation via probability map guided
  copy-pasting.
\newblock In {\em ICCV}, 2019.

\bibitem{georgakis2016multiview}
Georgios Georgakis, Md~Alimoor Reza, Arsalan Mousavian, Phi-Hung Le, and Jana
  Ko{\v{s}}eck{\'a}.
\newblock Multiview rgb-d dataset for object instance detection.
\newblock In {\em 3DV}, 2016.

\bibitem{ghiasi2019fpn}
Golnaz Ghiasi, Tsung-Yi Lin, and Quoc~V Le.
\newblock Nas-fpn: Learning scalable feature pyramid architecture for object
  detection.
\newblock In {\em CVPR}, 2019.

\bibitem{girshick2015fast}
Ross Girshick.
\newblock Fast r-cnn.
\newblock In {\em ICCV}, 2015.

\bibitem{girshick2014rich}
Ross Girshick, Jeff Donahue, Trevor Darrell, and Jitendra Malik.
\newblock Rich feature hierarchies for accurate object detection and semantic
  segmentation.
\newblock In {\em CVPR}, 2014.

\bibitem{girshick2018detectron}
Ross Girshick, Ilija Radosavovic, Georgia Gkioxari, Piotr Doll{\'a}r, and
  Kaiming He.
\newblock Detectron, 2018.

\bibitem{gupta2019lvis}
Agrim Gupta, Piotr Dollar, and Ross Girshick.
\newblock Lvis: A dataset for large vocabulary instance segmentation.
\newblock In {\em CVPR}, 2019.

\bibitem{hariharan2014simultaneous}
Bharath Hariharan, Pablo Arbel{\'a}ez, Ross Girshick, and Jitendra Malik.
\newblock Simultaneous detection and segmentation.
\newblock In {\em ECCV}, 2014.

\bibitem{hariharan2015hypercolumns}
Bharath Hariharan, Pablo Arbel{\'a}ez, Ross Girshick, and Jitendra Malik.
\newblock Hypercolumns for object segmentation and fine-grained localization.
\newblock In {\em CVPR}, 2015.

\bibitem{he2020momentum}
Kaiming He, Haoqi Fan, Yuxin Wu, Saining Xie, and Ross Girshick.
\newblock Momentum contrast for unsupervised visual representation learning.
\newblock In {\em CVPR}, 2020.

\bibitem{he2019rethinking}
Kaiming He, Ross Girshick, and Piotr Doll{\'a}r.
\newblock Rethinking imagenet pre-training.
\newblock In {\em ICCV}, 2019.

\bibitem{he2017mask}
Kaiming He, Georgia Gkioxari, Piotr Doll{\'a}r, and Ross Girshick.
\newblock Mask r-cnn.
\newblock In {\em ICCV}, 2017.

\bibitem{he2016deep}
Kaiming He, Xiangyu Zhang, Shaoqing Ren, and Jian Sun.
\newblock Deep residual learning for image recognition.
\newblock In {\em CVPR}, 2016.

\bibitem{henaff2019data}
Olivier~J H{\'e}naff, Aravind Srinivas, Jeffrey De~Fauw, Ali Razavi, Carl
  Doersch, SM Eslami, and Aaron van~den Oord.
\newblock Data-efficient image recognition with contrastive predictive coding.
\newblock {\em arXiv preprint arXiv:1905.09272}, 2019.

\bibitem{hu2020learning}
Xinting Hu, Yi Jiang, Kaihua Tang, Jingyuan Chen, Chunyan Miao, and Hanwang
  Zhang.
\newblock Learning to segment the tail.
\newblock In {\em CVPR}, 2020.

\bibitem{huang2016learning}
Chen Huang, Yining Li, Chen~Change Loy, and Xiaoou Tang.
\newblock Learning deep representation for imbalanced classification.
\newblock In {\em CVPR}, 2016.

\bibitem{ioffe2015batch}
Sergey Ioffe and Christian Szegedy.
\newblock Batch normalization: Accelerating deep network training by reducing
  internal covariate shift.
\newblock In {\em ICML}, 2015.

\bibitem{jamal2020rethinking}
Muhammad~Abdullah Jamal, Matthew Brown, Ming-Hsuan Yang, Liqiang Wang, and
  Boqing Gong.
\newblock Rethinking class-balanced methods for long-tailed visual recognition
  from a domain adaptation perspective.
\newblock In {\em CVPR}, 2020.

\bibitem{kang2020decoupling}
Bingyi Kang, Saining Xie, Marcus Rohrbach, Zhicheng Yan, Albert Gordo, Jiashi
  Feng, and Yannis Kalantidis.
\newblock Decoupling representation and classifier for long-tailed recognition.
\newblock In {\em ICLR}, 2020.

\bibitem{khan2019striking}
Salman Khan, Munawar Hayat, Syed~Waqas Zamir, Jianbing Shen, and Ling Shao.
\newblock Striking the right balance with uncertainty.
\newblock In {\em CVPR}, 2019.

\bibitem{krizhevsky2012}
Alex Krizhevsky, Ilya Sutskever, and Geoffrey~E Hinton.
\newblock Imagenet classification with deep convolutional neural networks.
\newblock In {\em NeurIPS}, 2012.

\bibitem{lecun98gradient}
Yann LeCun, L{\'e}on Bottou, Yoshua Bengio, and Patrick Haffner.
\newblock Gradient-based learning applied to document recognition.
\newblock {\em Proceedings of the IEEE}, 1998.

\bibitem{li2020overcoming}
Yu Li, Tao Wang, Bingyi Kang, Sheng Tang, Chunfeng Wang, Jintao Li, and Jiashi
  Feng.
\newblock Overcoming classifier imbalance for long-tail object detection with
  balanced group softmax.
\newblock In {\em CVPR}, 2020.

\bibitem{lin2017feature}
Tsung-Yi Lin, Piotr Doll{\'a}r, Ross Girshick, Kaiming He, Bharath Hariharan,
  and Serge Belongie.
\newblock Feature pyramid networks for object detection.
\newblock In {\em CVPR}, 2017.

\bibitem{lin2017focal}
Tsung-Yi Lin, Priya Goyal, Ross Girshick, Kaiming He, and Piotr Doll{\'a}r.
\newblock Focal loss for dense object detection.
\newblock In {\em ICCV}, 2017.

\bibitem{lin2014microsoft}
Tsung-Yi Lin, Michael Maire, Serge Belongie, James Hays, Pietro Perona, Deva
  Ramanan, Piotr Doll{\'a}r, and C~Lawrence Zitnick.
\newblock Microsoft coco: Common objects in context.
\newblock In {\em ECCV}, 2014.

\bibitem{mahajan2018exploring}
Dhruv Mahajan, Ross Girshick, Vignesh Ramanathan, Kaiming He, Manohar Paluri,
  Yixuan Li, Ashwin Bharambe, and Laurens van~der Maaten.
\newblock Exploring the limits of weakly supervised pretraining.
\newblock In {\em ECCV}, 2018.

\bibitem{oksuz2020imbalance}
Kemal Oksuz, Baris~Can Cam, Sinan Kalkan, and Emre Akbas.
\newblock Imbalance problems in object detection: A review.
\newblock {\em TPAMI}, 2020.

\bibitem{qiao2020detectors}
Siyuan Qiao, Liang-Chieh Chen, and Alan Yuille.
\newblock Detectors: Detecting objects with recursive feature pyramid and
  switchable atrous convolution.
\newblock {\em arXiv preprint arXiv:2006.02334}, 2020.

\bibitem{radosavovic2018data}
Ilija Radosavovic, Piotr Doll{\'a}r, Ross Girshick, Georgia Gkioxari, and
  Kaiming He.
\newblock Data distillation: Towards omni-supervised learning.
\newblock In {\em CVPR}, 2018.

\bibitem{remez2018learning}
Tal Remez, Jonathan Huang, and Matthew Brown.
\newblock Learning to segment via cut-and-paste.
\newblock In {\em ECCV}, 2018.

\bibitem{ren2020balanced}
Jiawei Ren, Cunjun Yu, Zhongang Cai, and Haiyu Zhao.
\newblock Balanced activation for long-tailed visual recognition.
\newblock In {\em LVIS Challenge Workshop at ECCV}, 2020.

\bibitem{ren2015faster}
Shaoqing Ren, Kaiming He, Ross Girshick, and Jian Sun.
\newblock Faster r-cnn: Towards real-time object detection with region proposal
  networks.
\newblock In {\em NeurIPS}, 2015.

\bibitem{russakovsky2015imagenet}
Olga Russakovsky, Jia Deng, Hao Su, Jonathan Krause, Sanjeev Satheesh, Sean Ma,
  Zhiheng Huang, Andrej Karpathy, Aditya Khosla, Michael Bernstein, et~al.
\newblock Imagenet large scale visual recognition challenge.
\newblock {\em IJCV}, 2015.

\bibitem{shao2019objects365}
Shuai Shao, Zeming Li, Tianyuan Zhang, Chao Peng, Gang Yu, Xiangyu Zhang, Jing
  Li, and Jian Sun.
\newblock Objects365: A large-scale, high-quality dataset for object detection.
\newblock In {\em ICCV}, 2019.

\bibitem{shorten2019survey}
Connor Shorten and Taghi~M Khoshgoftaar.
\newblock A survey on image data augmentation for deep learning.
\newblock {\em Journal of Big Data}, 2019.

\bibitem{simonyan2014very}
Karen Simonyan and Andrew Zisserman.
\newblock Very deep convolutional networks for large-scale image recognition.
\newblock In {\em ICLR}, 2015.

\bibitem{singh2018sniper}
Bharat Singh, Mahyar Najibi, and Larry~S Davis.
\newblock Sniper: Efficient multi-scale training.
\newblock In {\em NeurIPS}, 2018.

\bibitem{szegedy2015}
Christian Szegedy, Wei Liu, Yangqing Jia, Pierre Sermanet, Scott Reed, Dragomir
  Anguelov, Dumitru Erhan, Vincent Vanhoucke, and Andrew Rabinovich.
\newblock Going deeper with convolutions.
\newblock In {\em CVPR}, 2015.

\bibitem{tan2020equalization}
Jingru Tan, Changbao Wang, Buyu Li, Quanquan Li, Wanli Ouyang, Changqing Yin,
  and Junjie Yan.
\newblock Equalization loss for long-tailed object recognition.
\newblock In {\em CVPR}, 2020.

\bibitem{lviswinner}
Jingru Tan, Gang Zhang, Hanming Deng, Changbao Wang, Lewei Lu, Quanquan Li, and
  Jifeng Dai.
\newblock 1st place solution of lvis challenge 2020: A good box is not a
  guarantee of a good mask.
\newblock {\em arXiv preprint arXiv:2009.01559}, 2020.

\bibitem{tan2019efficientnet}
Mingxing Tan and Quoc~V Le.
\newblock Efficientnet: Rethinking model scaling for convolutional neural
  networks.
\newblock In {\em ICML}, 2019.

\bibitem{tan2019efficientdet}
Mingxing Tan, Ruoming Pang, and Quoc~V Le.
\newblock Efficientdet: Scalable and efficient object detection.
\newblock In {\em CVPR}, 2020.

\bibitem{tang2020long}
Kaihua Tang, Jianqiang Huang, and Hanwang Zhang.
\newblock Long-tailed classification by keeping the good and removing the bad
  momentum causal effect.
\newblock {\em NeurIPS}, 2020.

\bibitem{van2018inaturalist}
Grant Van~Horn, Oisin Mac~Aodha, Yang Song, Yin Cui, Chen Sun, Alex Shepard,
  Hartwig Adam, Pietro Perona, and Serge Belongie.
\newblock The inaturalist species classification and detection dataset.
\newblock In {\em CVPR}, 2018.

\bibitem{wan2013regularization}
Li Wan, Matthew Zeiler, Sixin Zhang, Yann Le~Cun, and Rob Fergus.
\newblock Regularization of neural networks using dropconnect.
\newblock In {\em ICML}, 2013.

\bibitem{wang2020scaled}
Chien-Yao Wang, Alexey Bochkovskiy, and Hong-Yuan~Mark Liao.
\newblock Scaled-yolov4: Scaling cross stage partial network.
\newblock {\em arXiv preprint arXiv:2011.08036}, 2020.

\bibitem{wang2020devil}
Tao Wang, Yu Li, Bingyi Kang, Junnan Li, Junhao Liew, Sheng Tang, Steven Hoi,
  and Jiashi Feng.
\newblock The devil is in classification: A simple framework for long-tail
  object detection and instance segmentation.
\newblock In {\em ECCV}, 2020.

\bibitem{wang2019meta}
Yu-Xiong Wang, Deva Ramanan, and Martial Hebert.
\newblock Meta-learning to detect rare objects.
\newblock In {\em ICCV}, 2019.

\bibitem{xie2020self}
Qizhe Xie, Minh-Thang Luong, Eduard Hovy, and Quoc~V Le.
\newblock Self-training with noisy student improves imagenet classification.
\newblock In {\em CVPR}, 2020.

\bibitem{yun2019cutmix}
Sangdoo Yun, Dongyoon Han, Seong~Joon Oh, Sanghyuk Chun, Junsuk Choe, and
  Youngjoon Yoo.
\newblock Cutmix: Regularization strategy to train strong classifiers with
  localizable features.
\newblock In {\em ICCV}, 2019.

\bibitem{zhang2017mixup}
Hongyi Zhang, Moustapha Cisse, Yann~N Dauphin, and David Lopez-Paz.
\newblock mixup: Beyond empirical risk minimization.
\newblock In {\em ICLR}, 2018.

\bibitem{zhang2020resnest}
Hang Zhang, Chongruo Wu, Zhongyue Zhang, Yi Zhu, Zhi Zhang, Haibin Lin, Yue
  Sun, Tong He, Jonas Mueller, R. Manmatha, Mu Li, and Alexander Smola.
\newblock Resnest: Split-attention networks.
\newblock In {\em arXiv preprint arXiv:2004.08955}, 2020.

\bibitem{zhang2018single}
Shifeng Zhang, Longyin Wen, Xiao Bian, Zhen Lei, and Stan~Z Li.
\newblock Single-shot refinement neural network for object detection.
\newblock In {\em CVPR}, 2018.

\bibitem{zhang2019bag}
Zhi Zhang, Tong He, Hang Zhang, Zhongyue Zhang, Junyuan Xie, and Mu Li.
\newblock Bag of freebies for training object detection neural networks.
\newblock {\em arXiv preprint arXiv:1902.04103}, 2019.

\bibitem{zhang2018exfuse}
Zhenli Zhang, Xiangyu Zhang, Chao Peng, Xiangyang Xue, and Jian Sun.
\newblock Exfuse: Enhancing feature fusion for semantic segmentation.
\newblock In {\em ECCV}, 2018.

\bibitem{zhou2020bbn}
Boyan Zhou, Quan Cui, Xiu-Shen Wei, and Zhao-Min Chen.
\newblock Bbn: Bilateral-branch network with cumulative learning for
  long-tailed visual recognition.
\newblock In {\em CVPR}, 2020.

\bibitem{zoph2019learning}
Barret Zoph, Ekin~D Cubuk, Golnaz Ghiasi, Tsung-Yi Lin, Jonathon Shlens, and
  Quoc~V Le.
\newblock Learning data augmentation strategies for object detection.
\newblock In {\em ECCV}, 2020.

\bibitem{zoph2020rethink}
Barret Zoph, Golnaz Ghiasi, Tsung-Yi Lin, Yin Cui, Hanxiao Liu, Ekin~D. Cubuk,
  and Quoc~V. Le.
\newblock Rethinking pre-training and self-training.
\newblock In {\em NeurIPS}, 2020.

\end{thebibliography}
